\documentclass[11pt]{article}

\usepackage{amssymb}
\usepackage{amsmath,amsfonts}
\usepackage{mathtools}
\usepackage{amsthm}
\usepackage{latexsym}
\usepackage{mathrsfs}
\usepackage{color}
\usepackage{comment}
\usepackage{hyperref}
\usepackage{graphicx}
\usepackage{booktabs}
\usepackage{authblk}
\usepackage{subcaption}
\usepackage[ruled,vlined,linesnumbered,inoutnumbered]{algorithm2e} 
\usepackage[a4paper]{geometry}
\usepackage[numbers]{natbib}
\usepackage{bbm}
\usepackage{multirow}
\usepackage[title]{appendix}

\newcommand{\argmin}{\arg\min}
\newcommand{\argmax}{\arg\max}
\DeclareMathOperator{\OneHot}{\sf OneHot}
\DeclareMathOperator{\MultiHot}{\sf MultiHot}
\DeclareMathOperator{\softmax}{\sf softmax}
\DeclareMathOperator{\supp}{\sf supp}
\DeclareMathOperator{\crossenrtopy}{\sf crossentropy}
\DeclareMathOperator{\diag}{\sf diag}

\newcommand{\Bscr}{\mathcal{B}}
\newcommand{\Cscr}{\mathcal{C}}
\newcommand{\Dscr}{\mathcal{D}}

\newcommand{\Fscr}{\mathcal{F}}

\newcommand{\Lscr}{\mathcal{L}}

\newcommand{\Pscr}{\mathcal{P}}

\newcommand{\Tscr}{\mathcal{T}}

\newcommand{\Xscr}{\mathcal{X}}

\newcommand{\mE}{\mathbb{E}}

\newcommand{\mR}{\mathbb{R}}

\newtheorem{assumption}{Assumption}[section]
\newtheorem{proposition}{Proposition}[section]
\newtheorem{definition}{Definition}[section]

\newtheorem{requirement}{Requirement}[section]

\definecolor{blu}{rgb}{0,0,1}

\definecolor{gre}{rgb}{0,.5,0}

\definecolor{red}{rgb}{1,0,0}

\usepackage{color, soul}

\begin{document}

\title{Knowledge-Injected Federated Learning}
\author[1,2]{Zhenan Fan}
\author[2]{Zirui Zhou}
\author[4]{Jian Pei}
\author[1]{Michael P. Friedlander}
\author[3]{Jiajie Hu}
\author[3]{Chengliang Li}
\author[2]{Yong Zhang}
\affil[1]{University of British Columbia, 
        \texttt{ \{zhenanf, mpf\}@cs.ubc.ca} }
\affil[2]{Huawei Technologies Canada, 
        \texttt{ \{zirui.zhou, yong.zhang3\}@huawei.com} }
\affil[3]{Huawei Technologies China, 
        \texttt{ \{hujiajie1, lichengliang2\}@huawei.com} }
\affil[4]{Simon Fraser University, 
        \texttt{ jpei@cs.sfu.ca} }
\date{\today}
\maketitle

\begin{abstract}
Federated learning is an emerging technique for training models from decentralized data sets. In many applications, data owners participating in the federated learning system hold not only the data but also a set of domain knowledge. Such knowledge includes human know-how and craftsmanship that can be extremely helpful to the federated learning task. In this work, we propose a federated learning framework that allows the injection of participants' domain knowledge, where the key idea is to refine the global model with knowledge locally. The scenario we consider is motivated by a real industry-level application, and we demonstrate the effectiveness of our approach to this application. 
\end{abstract}

\section{Introduction} \label{sec:intro}

With the development of artificial intelligence, people recognize that many powerful machine learning models are driven by large decentralized datasets of various data types. However, in many industry-scale applications, training data is obtained and maintained by different data owners instead of centralized at the data center, and sharing data is often forbidden due to privacy requirements. Federated learning (FL) is an emerging machine learning framework in which multiple data owners (also referred to as clients) participate in collaboratively training a model without sharing their local data with each other~\citep{federated2016,yang2019federated}.

Another challenge with artificial intelligence is integrating domain knowledge into purely data-driven models, i.e., parameters of the model are learned through training data without any human engineering~\citep{Chen2022IntegrationOK,E2021TheDO}. For example, human know-how and craftsmanship, which may not be learnable from the training data, can be formulated as prediction models, and combing them with a purely data-driven model may boost its performance and reduce the risk of overfitting~\cite{diez2019data}. As another example, knowledge of natural laws or security guidelines can be formulated as external constraints, and incorporating such constraints can prevent a purely data-driven model from making unreasonable predictions~\cite{borghesi2020improving}. These circumstances have led to increased research on improving machine learning models by additionally incorporating domain knowledge into the learning process~\cite{vonRueden2021InformedML}. 

\begin{figure}[t]
    \centering
    \includegraphics[width=.5\linewidth]{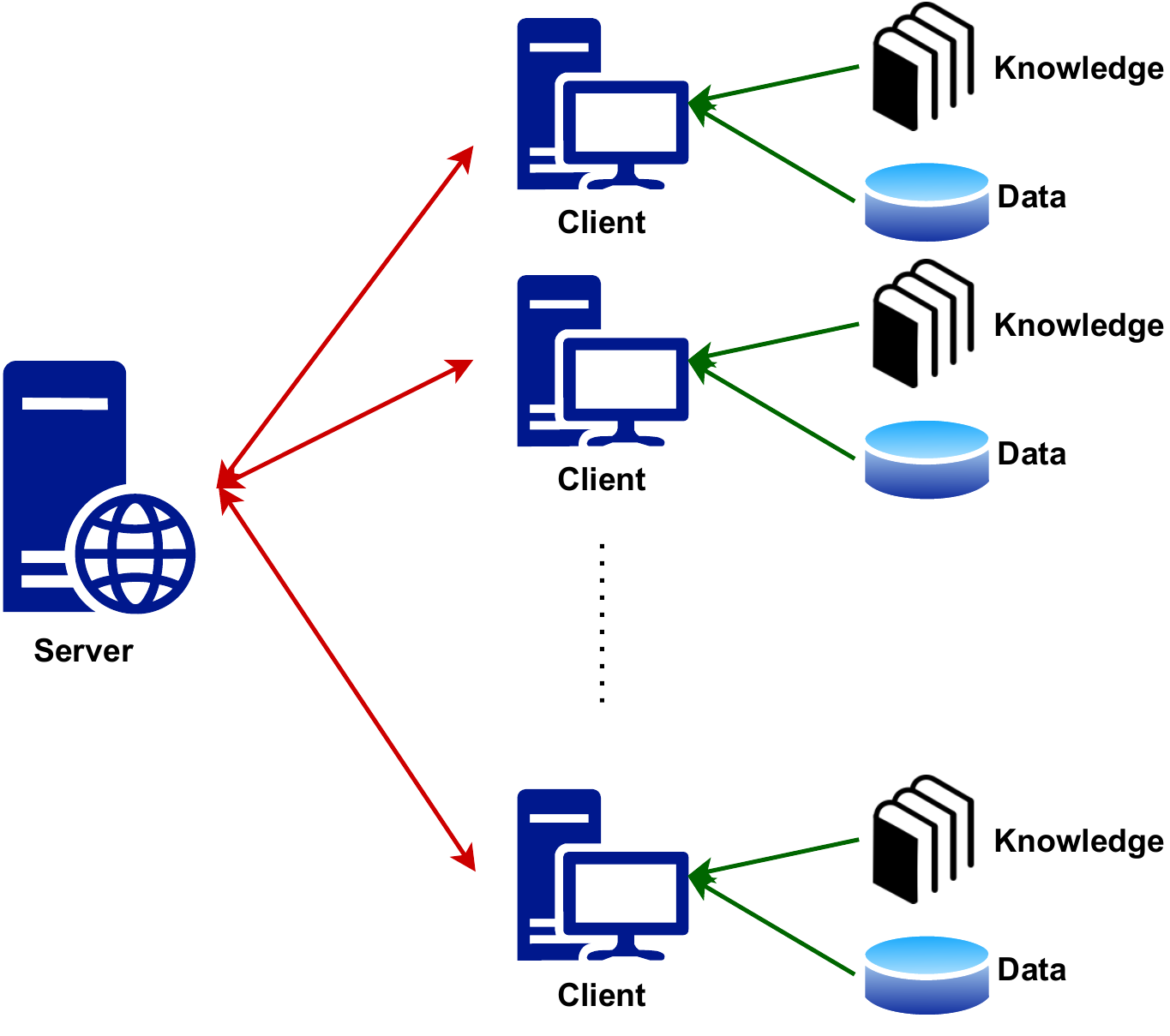}
    \caption{Knowledge-injected federated learning.}
    \label{fig:FedKnowledge}
\end{figure}

It is then natural to consider a situation where data owners have not only local data sets but also local domain knowledge. In Section~\ref{sec:case}, we show a real industrial application satisfying this situation, which also motivates this research work. Although local domain knowledge can be highly beneficial to the learning task, current techniques for FL tasks do not allow the injection of local knowledge. So it would be useful to provide a solution for FL that can account for and benefit from local knowledge and data sets. A simple illustration is shown in Figure~\ref{fig:FedKnowledge}.

In this work, we consider a collaborative learning task where there is a server responsible for invigilating the training process and multiple clients with different local data sets and domain knowledge. In particular, domain knowledge is represented in terms of knowledge-based models (KM). We consider two widely seen types of KMs: \emph{prediction-type knowledge-based model} (P-KM) and \emph{range-type knowledge-based model} (R-KM). The P-KM predicts a label for the given data instance, which is known to perform relatively well on the learning task, and the R-KM predicts a range of the possible labels for the given data instance such that the true label is guaranteed to be within this range. As mentioned above, P-KMs can represent human know-how and craftsmanship, and R-KMs can originate from natural laws or security guidelines. A more detailed introduction to KMs will be given in Section~\ref{sec:mech_model}.

We aim to design an FL framework such that the local domain knowledge can be injected. More specifically, each client is expected to obtain a local model that utilizes the local P-KM as a reference, and its prediction falls into the range given by the local R-KM. Moreover, as these KMs can be very valuable, their privacy should be protected as well as the local dataset. The more detailed setting and requirements will be given in Section~\ref{sec:setting}. 

Our contribution can be summarized as follows. We propose a framework for solving the knowledge-injected FL problem. Suppose the server provides the structure of a general deep learning model, which we call it server model. The key idea behind our framework is the design of a local function-transformation mapping that depends on the local KMs; see Definition~\ref{def:function_transform}. The function-transformation mapping can locally transform the server model into a personalized local model that satisfies the above-mentioned requirements; see Proposition~\ref{prop:property}. Moreover, we show that most existing FL algorithms can be applied to solve our proposed system and the privacy of local KMs can be protected in the sense that they are not shared during the training process; see Section~\ref{sec:algo}. Finally, we verify the effectiveness of our framework on a real industry-level problem (Section~\ref{sec:case}), and extensive numerical experiments on real and public data sets (Section~\ref{sec:experiments}).

\section{Related work} \label{sec:related}

According to our investigation, we haven't found any work on injecting domain knowledge into the FL process. So in this section, we instead review some literature on two related topics: knowledge-injected machine learning and personalized FL. 

\paragraph{Knowledge-injected machine learning.} Improving machine learning models via incorporating prior knowledge into the learning process has recently gained a lot of attention~\citep{von2019informed,borghesi2020improving}. We mainly focus on the works where the knowledge is interpreted as hard constraints on the prediction, which is similar to the role of the R-KM in our case. This task is challenging because a direct formulation would lead to an optimization problem with infinitely many constraints. To tackle this problem, \citet{hu2016harnessing} suggest incorporating the hard constraints with unlabelled data so that when the unlabelled data set is large enough, the final model is expected to satisfy the constraints. Similarly, \citet{nandwani2019primal} suggest relaxing the constraints to the training set and then solving the relaxed constrained optimization problem via a primal-dual method. When the training data set is large enough, the final model is also expected to satisfy the constraints. Alternatively, \citet{lee2019gradient} suggest incorporating the hard constraints into the inference process instead of the training process. They propose a gradient-based inference method, which can enforce the outputs of the model to satisfy the hard constraint. In this work, we instead suggest incorporating the hard constraints into the model formulation so that the output is guaranteed to satisfy the constraints. 

\paragraph{Personalized federated learning.} Personalized FL refers to the research on adjusting the FL process to solve the data heterogeneity issue~\citep{deng2020adaptive,fallah2020personalized}. Although personalization is not the motivation for this work, based on our proposed framework, every client receives a personalized model at the end of the training. Broadly speaking, personalization can be divided into two categories: (1) every client has the same model structure but with different weights, and (2) clients' models differ slightly in structure. There are many approaches that fall into the first category including local fine-tuning~\citep{mansour2020three,wang2019federated}, meta-learning~\citep{chen2018federated,jiang2019improving,fallah2020personalized} and regularization~\citep{t2020personalized,huang2021personalized}. There are relatively fewer works that belong to the second category. \citet{deng2020adaptive} propose to adapt the global model through a convex combination with clients' local models. Alternatively, \citet{arivazhagan2019federated} suggest clients locally add a personalized layer to the global model and the weights of the personalized layer will get updated during the training. Our approach can be seen as a combination of \citep{deng2020adaptive} and \citep{arivazhagan2019federated}, where the major difference is that in our case, the personalized layer consists of the local KMs and its parameters are fixed. We postpone a more detail discussion to Section~\ref{sec:architecture}.

\section{Preliminaries}

We introduce some notations used in this paper. We use $[k]$ to denote the set of integers $\{1,\dots,k\}$. We denote by $\OneHot(k)$ the set of $k$-dimensional binary vectors with exactly one entry being non-zero, and $\MultiHot(k)$ the set of $k$-dimensional binary vectors with at least one entry being non-zero. We use $\Delta^k$ to denote the $k$-dimensional simplex, i.e., 
\[\Delta^k = \left\{x \in \mR_+^k ~\big\vert~ \sum_{i=1}^k x_i = 1 \right\}.\]
For any $z\in\mR^k$, we use $\supp(z)$ to represent the set of nonzero entries of $z$, i.e., 
\[\supp(z) = \{i \mid z_i \neq 0\}.\]
The mapping $\softmax: \mR^k \to \Delta^k$ is defined by 
\[\softmax(z)_i = \frac{\exp(z_i)}{\sum_{j=1}^k \exp(z_j)} \enspace \forall i = 1,\dots,k.\]
The mapping $\crossenrtopy: \Delta^k \times \Delta^k \to \mR$ is defined by 
\[\crossenrtopy(p, q) = -\sum_{i=1}^k q_i\log(p_i).\]
We define $\circ$ as the composition operator, i.e., for any two mappings $f:\mR^{m} \to \mR^n$ and $g:\mR^k\to\mR^m$, we have 
\[f\circ g(x) = f(g(x)) \forall x \in \mR^k.\]
We let $\Cscr(\mR^n, \mR^k)$ be the set of mappings from $\mR^n$ to $\mR^k$.

\section{Knowledge-based models} \label{sec:mech_model} 
We consider the scenario where the knowledge held by each client of the FL system is in the form of KMs. These models provide guidance on predictions based on various types of knowledge, such as empirical evidence, mechanism of the application domain, and causality of the input-output relationship. Below we provide a formal definition of KMs in the context of classification.

Recall that in a standard $k$-class classification problem, we are given \emph{i.i.d.} data samples $\Dscr = \{(x^{(i)}, y^{(i)})\}_{i=1}^N \subset \Xscr\times\OneHot(k)$ that are drawn from an underlying distribution $\Fscr$
and a hypothesis space $\Theta$ that is a set of mappings from $\Xscr$ to $[k]$. The goal is to find a mapping $f^*\in\Theta$ such that
\begin{equation} \label{eq:multiclass}
    f^* = \argmin_{f\in\Theta} \mE_{(x,y)\sim\Fscr} [ f(x) \neq y].
\end{equation}

\begin{definition}\label{def:know-model}
We say that a mapping $g$ is a KM for the data distribution $\Fscr$ if it satisfies the following conditions:
\begin{itemize}
    \item[(a)] The domain of $g$ equals to the feature space $\Xscr$.
    \item[(b)] $g$ is independent of the training data $\Dscr$. 
    \item[(c)] For any $(x,y)\sim\Fscr$, $g(x)$ provides information on the true label $y$.
\end{itemize}
\end{definition}

Definition~\ref{def:know-model}(b) emphasizes that knowledge and training data are two independent sources of information on $\Fscr$, and Definition~\ref{def:know-model}(c) suggests that they can complement each other in solving the learning problem \eqref{eq:multiclass}. 

KMs can be divided into multiple categories depending on the type of information they provide. In this work, we focus on the following two types that are commonly seen in practice: \emph{prediction type} and \emph{range type}.

\begin{definition}\label{def:pred-model}
Suppose that $g$ is a KM for the data distribution $\Fscr$ (Definition~\ref{def:know-model}).
\begin{itemize}
\item[(a)] We say that $g$ is of prediction type if it is a mapping from $\Xscr$ to $\OneHot(k)$ and for any $(x,y)\sim\Fscr$, $g(x)$ is a point estimation of $y$.
\item[(b)] We say that $g$ is of range type if it is a mapping from $\Xscr$ to $\MultiHot(k)$ and for any $(x,y)\sim\Fscr$, one has
$$ \supp(y) \subseteq \supp(g(x)). $$
\end{itemize}
\end{definition}
In other words, for any $(x,y)\sim\Fscr$, a P-KM $g_p$ predicts a class label $g_p(x)$, which is an estimation of its true label $y$; while a R-KM $g_r$ states that the true label $y$ must be within $g_r(x)$, which is a subset of all the $k$ classes. Below are some motivating examples of the two models:
\begin{itemize}
\item[(a)] Mechanistic prediction models, such as differential equations that describe the underlying physical processes, can be considered as P-KMs. These models normally generalize well. On the other hand, since they are usually obtained by simplifying the real-world process and ignoring unpredictable factors, their accuracy can be improved by combining them with data-driven approaches.
\item[(b)] Prediction models learned in the past can also be P-KMs for the new learning task. Suppose that a prediction model $g$ had been learned from data collected before the equipment was recently updated and a set $\Dscr$ of new data samples has been collected since the update. As the updated equipment may use more advanced technology, such as cameras with higher resolution and wearable devices with more accurate sensors, the new data set $\Dscr$ has the potential to lead to a more accurate prediction model than $g$. Besides, since $g$ was learned from a much larger set of data than $\Dscr$ and had been continuously refined in practice, it can be helpful to prevent the newly trained model from overfitting. 
\item[(c)] A R-KM can be derived from the causality of the input-output relationship. In practice, domain experts usually have some causality-based rules that any prediction model has to satisfy. For example, if the value of some feature of a data sample is above a certain threshold, then its label must be within some subset of $[k]$. R-KMs can be injected into the training process to prevent the learned model from making predictions of no sense.
\end{itemize}

\section{Federated learning with knowledge-based models} \label{sec:methodology}
This section formally introduces the collaborative learning task and our proposed methodology. 

\subsection{Problem setting} \label{sec:setting} 
Suppose there are $M$ clients. Each client $m$ has a training data set $\Dscr^m \sim \Fscr^m$, where $\Fscr^m$ is a conditional data distribution depending on underlying data distribution $\Fscr$. For example, $\Fscr^m$ can be the conditional distribution such that the label is fixed to be $1$, i.e., 
\[\Fscr^m = \{(x,y) \sim \Fscr \mid \supp(y) = \{1\} \}.\]
Moreover, each client $m$ has a P-KM $g_p^m$ and a R-KM $g_r^m$ for the local data distribution $\Fscr^m$ (Definition~\ref{def:pred-model}). Besides the clients, there is also a server that designs the global model structure and organizes the training. We will describe the roles of the server in Section~\ref{sec:architecture} and Section~\ref{sec:algo}. 

As we introduced in Section~\ref{sec:mech_model}, local training data and KMs are two different sources for data distribution. So integrating the local KMs into the learning process can benefit this collaborative learning task. Moreover, similar to data, KMs are assets for their owners. Indeed, it may be obtained via years of investment, and it can be valuable to its owner's business. So their privacy should be protected during the training process. 

More specifically, we want to design an FL framework so that each client $m$ obtains a personalized, predictive model $f^m: \Xscr \to \Delta^k$ satisfying some requirements listed below. 
\begin{requirement} \label{requirements}
    We want the framework to satisfy the following requirements. 
    \begin{itemize}
    \item[(a)] Each personalized model $f^m$ utilizes the local P-KM $g_p^m$ where the trust level of $g_p^m$ is controllable;
    \item[(b)] Each personalized model $f^m$ agrees with the local R-KM $g_r^m$, i.e., for any $x \in \Xscr$, we have 
    \[\supp(f^m(x)) \subseteq \supp(g_r^m(x));\]
    \item[(c)] Privacy of local dataset and local KMs needs to be protected.
\end{itemize}
\end{requirement}

The following assumption guarantees that there is no conflict between local P-KMs and R-KMs. 
\begin{assumption} \label{ass:blanket}
    For each client $m$, the local P-KM $g_p^m$ agrees with the local R-KM $g_r^m$, i.e., for any $x \in \Xscr$, we have
    \[\supp(g_p^m(x)) \subseteq \supp(g_r^m(x)).\]
\end{assumption}

\subsection{Architecture design} \label{sec:architecture}

Now we show how to design the personalized model $f^m$. Suppose the server provides a general deep learning model $f(\theta; \cdot):\Xscr\to\mR^k$ with $\theta \in \mR^d$ being the model parameters, e.g. multilayer perceptron network and convolutional neural network, which we call \emph{server model}. Our key idea is to design a function-transformation mapping that can locally transform the server model $f$ into a local model $f^m$, which utilizes local KMs. We give an explicit formula for this function-transformation mapping in the following definition. 

\begin{definition}[Function transformation] \label{def:function_transform}
    Given a scalar $\lambda\in[0,1]$ and functions $g_p:\Xscr\to\OneHot(k)$ and $g_r:\Xscr\to\MultiHot(k)$, we develop a function-transformation mapping $\Tscr_{\lambda,g_p, g_r}:\Cscr(\Xscr, \mR^k) \to \Cscr(\Xscr, \mR^k)$ such that for all $x \in \Xscr$
    \begin{equation}
            \Tscr_{\lambda,g_p, g_r}(f)(x) 
            =  (1-\lambda)\cdot\softmax( f(x) + \infty\cdot(g_r(x) - \mathbf{1}^k)) + \lambda\cdot g_p(x),
    \end{equation}
    where $\mathbf{1}^k$ denotes the k-dimensional vector with all ones. 
\end{definition}

\begin{figure}[t]
    \centering
    \includegraphics[width=.5\linewidth]{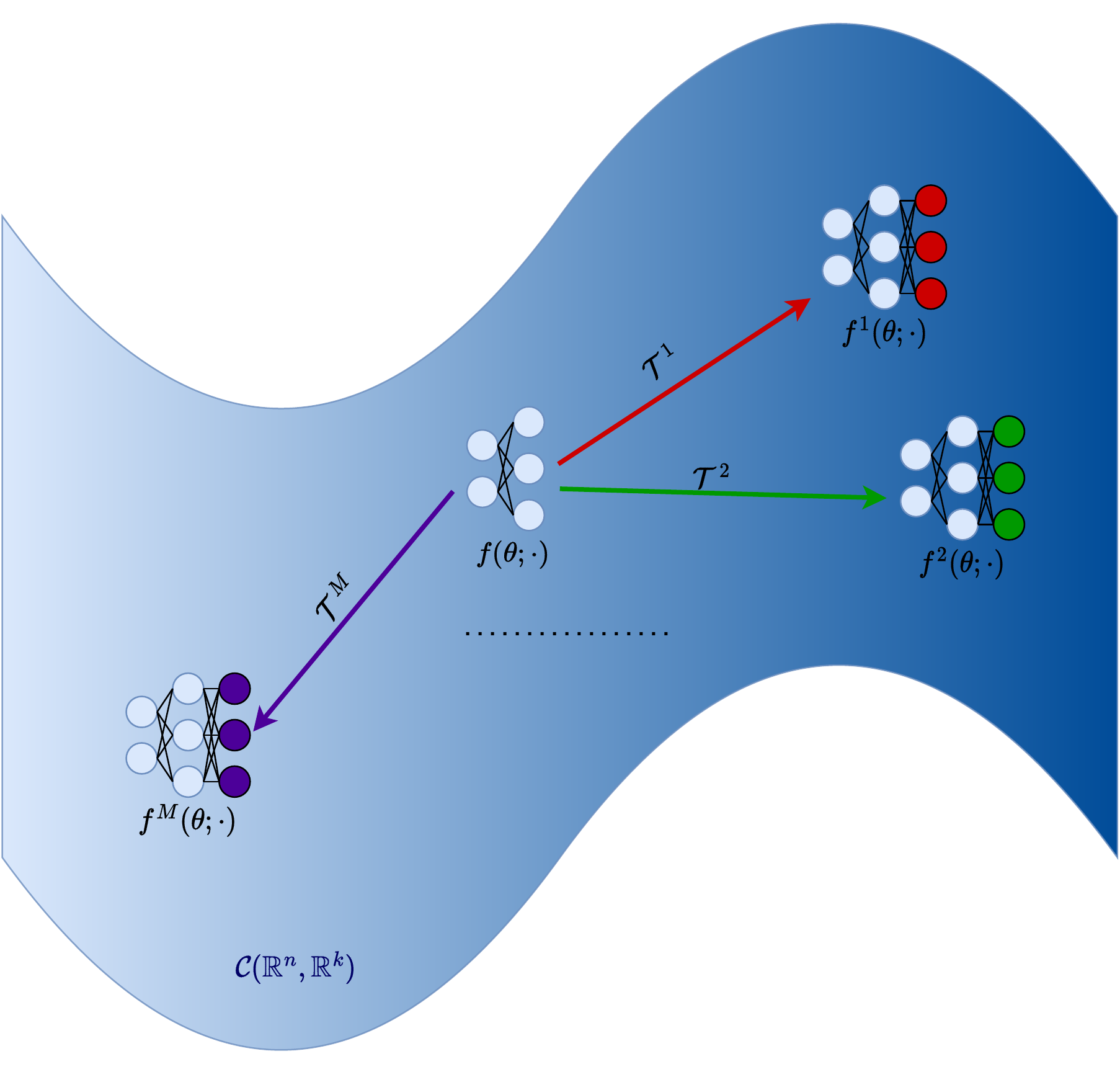}
    \caption{Illustration of the function transformation mapping $\Tscr_{\lambda,g_p,g_r}$ (Definition~\ref{def:function_transform}).}
    \label{fig:function_transformation}
\end{figure}

Based on the function-transformation mapping $\Tscr$, we can construct the local personalized model as
\begin{equation} \label{eq:local_model}
    f^m(\theta; \cdot) \coloneqq \Tscr^m(f(\theta; \cdot)) \enspace\text{with}\enspace \Tscr^m = \Tscr_{\lambda^m,g_p^m,g_r^m}.
\end{equation}
As we can see from the construction, all the personalized models $f^m$ share the same component $f$ with $\theta$ being the trainable global parameter, and the function-transformation mapping $\Tscr^m$ locally adds a layer consisting of local KMs $g_p^m$ and $g_r^m$, to $f$. A geometric illustration is shown is Figure~\ref{fig:function_transformation}. 

As we briefly discussed in Section~\ref{sec:intro}, both convex combination with local models~\citep{deng2020adaptive} and adding personalized layers~\citep{arivazhagan2019federated} are not new ideas in FL. In our design, both the structure and the parameters of the personalized layer are fixed and determined by the local KMs. The most important feature of our design is that the personalized model $f^m$ indeed satisfies the requirements we introduced in Section~\ref{sec:setting}, which will be shown in the following proposition. 

\begin{proposition}[Properties of the personalized model $f^m$] \label{prop:property}
    For any client $m$, let $f^m$ denote the personalized model constructed via \eqref{eq:local_model}. Under Assumption~\ref{ass:blanket}, the following three properties hold for $f^m(\theta; \cdot)$ with any model parameter $\theta \in \mR^d$. 
    \begin{enumerate}
        \item[(a)] $f^m(\theta; \cdot)$ is a valid predictive model in the sense that for any $x \in \Xscr$, we have
        \[f^m(\theta; x) \in \Delta^k. \] 
        \item[(b)] The scalar $\lambda^m \in [0, 1]$ controls the trust-level of the local P-KM $g_p^m$ in sense that for any $x\in\Xscr$, we have
        \[\langle f^m(\theta; x),\ g_p^m(x)\rangle \geq \lambda^m.\]
        Moreover, if $\lambda^m > 0.5$, then $f^m$ coincides with $g_p^m$ in the sense that for any $x\in\Xscr$, we have
        \[\argmax_i \{z_i \mid z = f^m(\theta; x)\} = \supp(g_p^m(x)).\]
        \item[(c)] $f^m(\theta; \cdot)$ agrees with the local R-KM $g_r^m$ for any parameter $\lambda^m\in[0,1]$ in the sense that for any $x\in\Xscr$, we have 
        \[\supp(f^m(\theta; x)) \subseteq \supp(g_r^m(x)).\]
        \item[(d)] Let $x\in\Xscr$ denote an arbitrary data point and let $\nabla_{\theta}f(\theta; x) \in \mR^{k\times d}$ denote the Jacobian matrix of the deep learning model $f(\theta; x)$. Then the Jacobian matrix of $f^m(\theta; x)$ can be expressed as 
        \[
            \nabla_{\theta}f^m(\theta; x) = (1-\lambda^m)\left[\diag(s) - s^T s\right]\cdot\nabla_{\theta}f(\theta; x),
        \]
        where
        \[s = \softmax( f(\theta; x) + \infty\cdot(h^m\circ\Phi^m(x) - \mathbbm{1}^k) ).\]
    \end{enumerate}
\end{proposition}

Proof of Proposition~\ref{prop:property} is contained in Appendix A. 
Proposition~\ref{prop:property} shows some nice properties of the personalized model $f^m$. More specifically, Proposition~\ref{prop:property}(a) guarantees that $f^m$ is a valid prediction model, Proposition~\ref{prop:property}(b) illustrates how $\lambda^m$ controls the trust-level of the local P-KM $g_p^m$ and shows that our design satisfies Requirement~\ref{requirements}(a), Proposition~\ref{prop:property}(c) shows that our design satisfies Requirement~\ref{requirements}(b), and Proposition~\ref{prop:property}(d) shows that the first order information of $f^m$ can be easily obtained as long as we know the Jacobian for the server model $f$. 

\subsection{Federated optimization} \label{sec:algo}
This section defines the optimization problem and shows how to update the global model parameter $\theta$. 

\paragraph{Optimization problem} Given the construction of the personalized models $f^m$ \eqref{eq:local_model}, we propose the following distributed empirical risk minimization problem
\begin{equation} \label{eq:fed_opt}
        \min_{\theta} \enspace \Lscr(\theta) \coloneqq \sum_{m=1}^M \Lscr^m(\theta)
\end{equation}
with 
\begin{equation} \label{eq:fed_opt_local}
    \Lscr^m(\theta) \coloneqq \frac{1}{|\Dscr^m|} \sum_{(x,y) \in \Dscr^m} \crossenrtopy( f^m(\theta; x), y ),
\end{equation}
where $\Lscr^m$ denotes the local empirical loss when applying personalized model $f^m$ to local dataset $\Dscr^m$. 

\paragraph{First order oracle} One advantage of our proposed model is the easy accessibility of its first order information. Assume that we have an oracle for computing the Jacobian matrix $\nabla_\theta f(\theta; x) \in \mR^{k \times d}$ for any data point $x \in \mR^n$, then given a batch $B^m \subset \Dscr^m$, the batch gradient of the local loss $\Lscr^m$ can be expressed as 
\begin{equation} \label{eq:batch_gradient}
        \nabla^{B^m}_\theta \Lscr^m(\theta) 
        = \frac{-1}{|B^m|} \sum_{(x,y) \in B^m} y^T\diag(f^m(\theta; x))^{-1} \nabla_\theta f^m(\theta; x),
\end{equation}
where the Jacobian matrix $\nabla_\theta f^m(\theta; x) \in \mR^{k \times d}$ can be computed as in Proposition~\ref{prop:property}(d).

\paragraph{Algorithm} Another advantage of our proposed model is that it is robust to optimization algorithms. Almost all the existing FL algorithms can be applied to solve \eqref{eq:fed_opt}, e.g. FedAvg\citep{mcmahan2017communication} SCAFFOLD~\citep{pmlr-v119-karimireddy20a} and Scaffnew~\citep{mishchenko2022proxskip}. As an example, we show how to apply FedAvg in Algorithm 1. 

\begin{algorithm}[t]
  \DontPrintSemicolon\setcounter{AlgoLine}{-1}
  \SetKwComment{tcp}{[}{]}
  \KwIn{server; $M$ clients; model initialization $\theta \in \mR^d$; sampling rate $\tau \in [0, 1]$; number of communication rounds $T$; batch size $C$; number of epochs in each round $E$; learning rate $\eta$} 
  \For{each round $t=0,1,\ldots,T$}{
        Server randomly samples a subset of clients $S_t \subset [M]$ with $|S_t| = \lceil\tau M\rceil$\;
        \For{each selected client $m \in S_t$ in parallel}{
            Client $m$ downloads latest model from server $\theta^m \leftarrow \theta$\;
            Client $m$ randomly splits local datasets $\Dscr^m$ into batches $\Bscr^m$ with batch size $C$\; 
            \For{each local epoch $e = 0, 1, \ldots, E$}{
                \For{each batch $B^m \in \Bscr^m_t$}{
                    Client $m$ computes local batch gradient $\nabla^{B^m}_\theta \Lscr^m(\theta^m)$ as in \eqref{eq:batch_gradient}\;
                    Client $m$ updates local model as $\theta^m \leftarrow \theta^m - \eta \nabla^{B^m}_\theta \Lscr^m(\theta^m)$\;
                }
            }
            Client $m$ uploads local model $\theta^m$ to server\;
        }
        Server updates global model $\theta \leftarrow \frac{1}{|S_t|}\sum_{m \in S_t} \theta^m$
  }
  \Return{$\theta$}
  \caption{\texttt{FedAvg} algorithm for solving the distributed optimization problem \eqref{eq:fed_opt}.
  \label{algo:fedavg}}
\end{algorithm}

\paragraph{Privacy}
As we can see from Algorithm 1, the only information passed between the server and the clients is the batch gradient, which is the same setting as in the conventional FL framework. As we can see from the expression~\eqref{eq:batch_gradient}, the batch gradient doesn't reveal the explicit formulas of the local KMs. Moreover, there has been an emerging paradigm in designing differentially private FL framework~\citep{hao2019towards,wei2020federated}. In particular, \citet{hao2019towards} show that differential privacy of the gradient mapping can be achieved by adding homomorphic encryption to the batch gradient. Their approach can be directly applied in our case because our proposed model is robust to optimization algorithms. 

\begin{figure}[t]
    \centering
    \includegraphics[width=.7\linewidth]{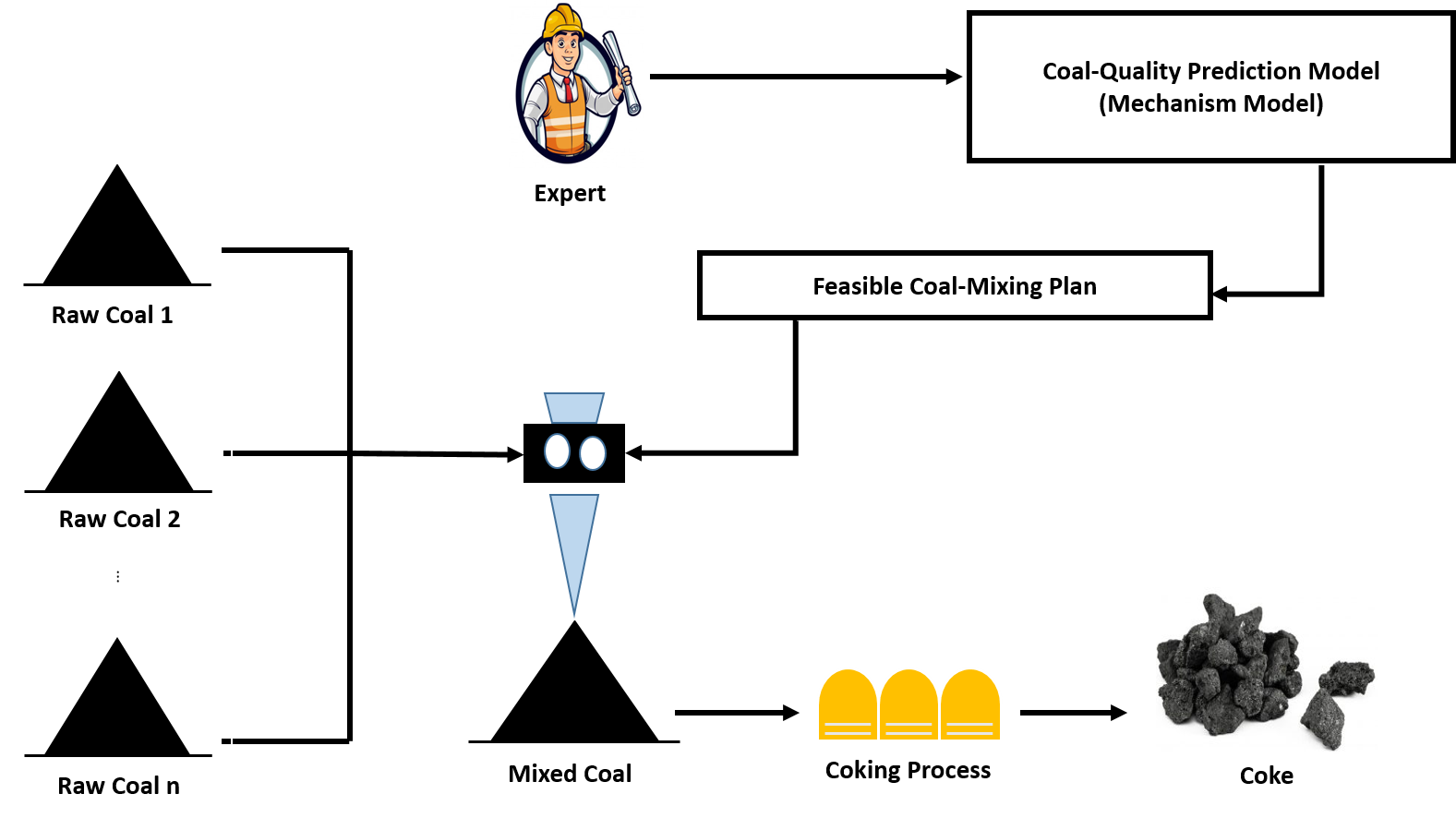}
    \caption{Illustration of coking process.}
    \label{fig:coke_process}
\end{figure}

\section{Case study: coal-mixing in coking process} \label{sec:case}

In this section, we conduct a case study on a real industry-level application that satisfies and motivates our framework. 

In China, coke is the main material in blast furnace iron-making, and its quality directly affects the output of steel. As a typical energy conversion industry, the coking industry plays a very important role in the national economy. Nearly 90\% of the coke it produces is used for iron smelting in the iron and steel industry, which greatly contributes to the country's industrialization, modernization, and urbanization process.

Due to the limited resources of high-quality coking coal, coking industries usually need to mix a variety of raw coal with a certain proportion. It is worth noting that the coal-mixing step accounts for up to 80\% of the key factors affecting the cost of the coking process. 

Determining the proportion of different raw coal is the key ingredient in the coal-mixing step, as it directly affects the quality of the final coke. It is also a very challenging task as there is no explicit formula modelling the relationship between the features of raw coal and the quality of the final coke. Currently, most domestic coking industries require technical experts to predict the quality of the final coke given a type of raw coal, which relies on the long-term experience of the experts. More specifically, technical experts will provide a P-KM and an R-KM, such that given features of a type of raw coal, the P-KM and the R-KM will, respectively, return a prediction and a range on the quality of the final coke. Finally, these KMs can provide an effective reference for engineers to obtain high-quality coke. Namely, given the requirements of coke quality, the engineers can find a feasible coal-mixing plan with the lowest cost, which can greatly improve production profits. The whole process is illustrated in Figure~\ref{fig:coke_process}.

However, due to the limitation of manual computing, technical experts usually can only utilize part of the features of the raw coal. As a consequence, the prediction given by the P-KM may be unstable, and although the range given by the R-KM will contain the true quality, it may be too large to be useful. This motivates us to use machine learning techniques to enhance the KMs so that the performance on the coke-quality prediction task can be improved. 

One bottleneck of applying machine learning techniques in this task is the scarcity of data because measuring the quality of the final coke is very expensive. To tackle this problem, we unite several coking industries to work on this task collaboratively, but their local datasets and KMs' privacy needs to be protected. 

Now we formally define the problem and present the numerical results. We have $M = 4$ coking industries. Each industry $m$ has a dataset $\Dscr^m = \{(x_i \in \mR^{17}, y_i \in \OneHot(20))\}_{i=1}^{N_m}$, where $x_i$ contains $17$ different features of the raw coal $i$ including its ash content, volatile fraction, coal rock reflectivity, etc. and $y_i$ presents the 20 grades of quality of coke made from the raw coal $i$. As we illustrated before, the local KMs only utilize part of the features due to the limitation of manual computing, i.e., each industry $m$ has a prediction model $\hat g_p^m: \mR^5 \to \OneHot(20)$ and a range model $\hat g_r^m: \mR^5 \to \OneHot(20)$. To match the definition of KMs (Definition~\ref{def:know-model}), we introduce a mask mapping $\Phi:\mR^{17}\to\mR^5$ and define the local P-KM and R-KM respectively as $g_r^m \equiv \hat g_p^m \circ \Phi$ and $g_r^m \equiv \hat g_p^m \circ \Phi$. The formal definition of the mask mapping $\Phi$ is given in Appendix B. Due to the company's regulations, we do not make the datasets public nor reveal the explicit formula of the KMs. We will test the performance of our approach on public datasets in Section~\ref{sec:experiments}. 

\begin{table}[t]
\centering
\begin{tabular}{|c|ccccc|ccccc|}
\hline
           & \multicolumn{5}{c|}{TA}         & \multicolumn{5}{c|}{POV}        \\ \hline
Industry   & ML   & P-KM & MLwKM   & FL   & FLwKM       & ML  & P-KM & MLwKM   & FL   & FLwKM   \\
1          & 56\% & 51\% & 66\%    & 68\% & 68\%        & 8\% & 0\%  & 0\%     & 7\%  & 0\%     \\
2          & 43\% & 40\% & 51\%    & 62\% & 63\%        & 16\% & 0\%  & 0\%     & 11\% & 0\%     \\
3          & 51\% & 38\% & 47\%    & 60\% & 61\%        & 7\% & 0\%  & 0\%     & 3\%  & 0\%     \\
4          & 42\% & 43\% & 52\%    & 67\% & 64\%        & 6\% & 0\%  & 0\%     & 12\% & 0\%     \\ \hline
\end{tabular}
\caption{Numerical results for the case study (Section~\ref{sec:case}). Comparison between the performance of five approaches introduced in Section~\ref{sec:case_study_numeric}. \label{tab:case_study}}
\vspace{-5mm}
\end{table}

\subsection{Numerical results} \label{sec:case_study_numeric}
In this case study, we compare the performance of five different approaches:
\begin{enumerate}
    \item \textbf{Machine learning (ML).} Each industry $m$ trains a model on its local training data set without the injection of local KMs and without FL. 
    \item \textbf{P-KM.} Each industry $m$ uses its local P-KM $g_p^m$ to make prediction.
    \item \textbf{Machine learning with KM (MLwKM).} Each industry $m$ locally trains the knowledge-injected model, i.e., minimizing $\Lscr^m(\theta)$ to obtain a local model $f^m$, where $\Lscr^m$ is defined in \eqref{eq:fed_opt_local}. 
    \item \textbf{Federated learning (FL).} All the industries conduct a standard FL without the injection of domain knowledge.
    \item \textbf{Federated learning with KM (FLwKM).} The approach we proposed in Section~\ref{sec:methodology}.
\end{enumerate}

We use two metrics to evaluate the performance of these approaches on the test data set. The first is the test accuracy (TA) and the second is the percentage of violation (POV) concerning the R-KM, i.e.
\begin{equation} \label{eq:per_vio}
    POV = \frac{1}{|\Dscr^m_{test}|} \sum_{(x,y) \in \Dscr^m_{test}} \mathbbm{1}(\{f^m(x) \notin h(x)\}).
\end{equation}

The numerical results are shown in Table~\ref{tab:case_study}. From the table, we can see that for industries 1, 2 and 3, FLwKM outperforms the other four approaches. For industry 4, although FL gives higher test accuracy, it has more violations than FLwKM. We postpone a more detailed comparison of these approaches to Section~\ref{sec:exp_approach}.

\section{Experiments} \label{sec:experiments}

In this section, we conduct experiments on real-world datasets to evaluate the effectiveness of our approach. We want to answer the following three questions: (i) Does our proposed approach help learn more powerful models in the low data setting? (Section~\ref{sec:exp_data}) (ii) Do KMs and the FL framework help learn more powerful models? (Section~\ref{sec:exp_approach}) (iii) What is the impact of hyperparameter $\lambda$? (Section~\ref{sec:exp_lambda})

We implement our approach in the Julia language~\cite{bezanson2017julia}. Our code is publicly available at \url{https://github.com/ZhenanFanUBC/FedMech.jl}.

\subsection{Data sets and distribution}

\paragraph{Covtype} The first dataset we use is the Forest Cover Type dataset \cite{blackard1999comparative}, where the task is to predict forest cover type from cartographic variables. The dataset is downloaded from the website of LIBSVM\footnote{https://www.csie.ntu.edu.tw/~cjlin/libsvm/}. It is a multi-class classification problem with the number of classes $k=7$ and the feature space $\Xscr = \mR^{54}$.

\paragraph{FMNIST} The second dataset we use is the Fashion MNIST dataset \cite{xiao2017fashion}, where the task is to predict the categories for grayscale
images of fashion products. The dataset is obtained from the Julia package \texttt{MLDatasets.jl}\footnote{https://github.com/JuliaML/MLDatasets.jl}. It is a multi-class classification problem with the number of classes $k=10$ and the feature space $\Xscr = \mR^{28\times 28}$.

\paragraph{Data distribution} For both datasets, we set the number of clients to be $5$ and distribute the data to clients in a non-i.i.d. fashion, i.e.,  each client gets samples of only $5$ classes, and the numbers of local training samples are not equal, which is the similar setting as in the FedAvg paper~\cite{mcmahan2017communication}. 

\subsection{Server and knowledge-based models}
                                                    
\paragraph{Server model $f(\theta, \cdot)$} For Covtype dataset, we set the server model $f(\theta, \cdot)$ to be the multi-layer perception model with two hidden layers. For FMNIST dataset, we set the server to be the famous LeNet-5 model originally proposed by \citet{lecun1998gradient}.     
                                        
\paragraph{P-KM $g_p^m$} For the Covtype dataset, we train a multinomial logistic regression model as the P-KM $g_p^m$ for each client $m$, where we use part of the training points with 18 features. Similar to Section~\ref{sec:case}, we use a mask mapping to model the limited feature situation. 
For the FMNIST dataset, we train a convolutional neural network as the P-KM $g_p^m$ for each client $m$, where we use part of the training images with a low-resolution view. We use a maxpooling operator to model the low-resolution situation. 

\paragraph{R-KM $g_r^m$} For both datasets, we locally construct a hashmap as the R-KM $g_r^m$ for each client $m$. Importantly, we guarantee that both the true label and the label predicted by the P-KM $g^m$ are contained in the range given by $g_r^m$, which satisfies the Assumption~\ref{ass:blanket}. 

\begin{table}[t]
\centering
\begin{tabular}{|c|cccc|cccc|}
\hline
       & \multicolumn{4}{c|}{Covtype}                & \multicolumn{4}{c|}{FMNIST}                 \\ \hline
Client & 1\% Data & 5\% Data & 10\% Data & 30\% Data & 1\% Data & 5\% Data & 10\% Data & 30\% Data \\ 
1      & 92\%     & 94\%     & 94\%      & 95\%      & 88\%     & 90\%     & 92\%      & 94\%      \\
2      & 71\%     & 83\%     & 88\%      & 90\%      & 79\%     & 87\%     & 87\%      & 90\%      \\
3      & 93\%     & 94\%     & 94\%      & 95\%      & 84\%     & 90\%     & 90\%      & 94\%      \\
4      & 79\%     & 83\%     & 84\%      & 86\%      & 84\%     & 89\%     & 91\%      & 93\%      \\
5      & 94\%     & 95\%     & 96\%      & 97\%      & 89\%     & 91\%     & 92\%      & 95\%      \\ \hline
\end{tabular}
\caption{Numerical results for the impact of the number of data, where the entries show the accuracy on the local test datasets (Section~\ref{sec:exp_data}). \label{tab:num_data}}
\end{table}

\begin{table}[t]
\centering
\begin{tabular}{|c|c|ccccc|ccccc|}
\hline
                   &        & \multicolumn{5}{c|}{Covtype}       & \multicolumn{5}{c|}{FMNIST}        \\ \hline
Client             & Metric & ML   & P-KM & MLwKM & FL   & FLwKM & ML   & P-KM & MLwKM & FL   & FLwKM \\ \hline
\multirow{2}{*}{1} & TA     & 87\% & 83\% & 89\%  & 90\% & 92\%  & 81\% & 78\% & 85\%  & 83\% & 88\%  \\
                   & POV    & 2\%  & 0\%  & 0\%   & 3\%  & 0\%   & 9\%  & 0\%  & 0\%   & 9\%  & 0\%   \\ \hline
\multirow{2}{*}{2} & TA     & 54\% & 59\% & 62\%  & 65\% & 71\%  & 74\% & 59\% & 69\%  & 74\% & 79\%  \\
                   & POV    & 10\% & 0\%  & 0\%   & 10\% & 0\%   & 12\% & 0\%  & 0\%   & 12\% & 0\%   \\ \hline
\multirow{2}{*}{3} & TA     & 87\% & 83\% & 90\%  & 88\% & 93\%  & 82\% & 61\% & 80\%  & 82\% & 84\%  \\
                   & POV    & 4\%  & 0\%  & 0\%   & 6\%  & 0\%   & 8\%  & 0\%  & 0\%   & 8\%  & 0\%   \\ \hline
\multirow{2}{*}{4} & TA     & 65\% & 67\% & 79\%  & 67\% & 79\%  & 76\% & 65\% & 76\%  & 76\% & 84\%  \\
                   & POV    & 15\% & 0\%  & 0\%   & 13\% & 0\%   & 12\% & 0\%  & 0\%   & 12\% & 0\%   \\ \hline
\multirow{2}{*}{5} & TA     & 92\% & 87\% & 92\%  & 92\% & 94\%  & 85\% & 79\% & 85\%  & 85\% & 89\%  \\
                   & POV    & 2\%  & 0\%  & 0\%   & 3\%  & 0\%   & 6\%  & 0\%  & 0\%   & 6\%  & 0\%   \\ \hline
\end{tabular}
\caption{Numerical results for the impact of KMs and the FL framework, where the entries show the accuracy on the local test datasets (Section~\ref{sec:exp_approach}). \label{tab:approach}}
\vspace{-5mm}
\end{table}

\begin{figure}[t]
\centering
\includegraphics[width=.7\linewidth]{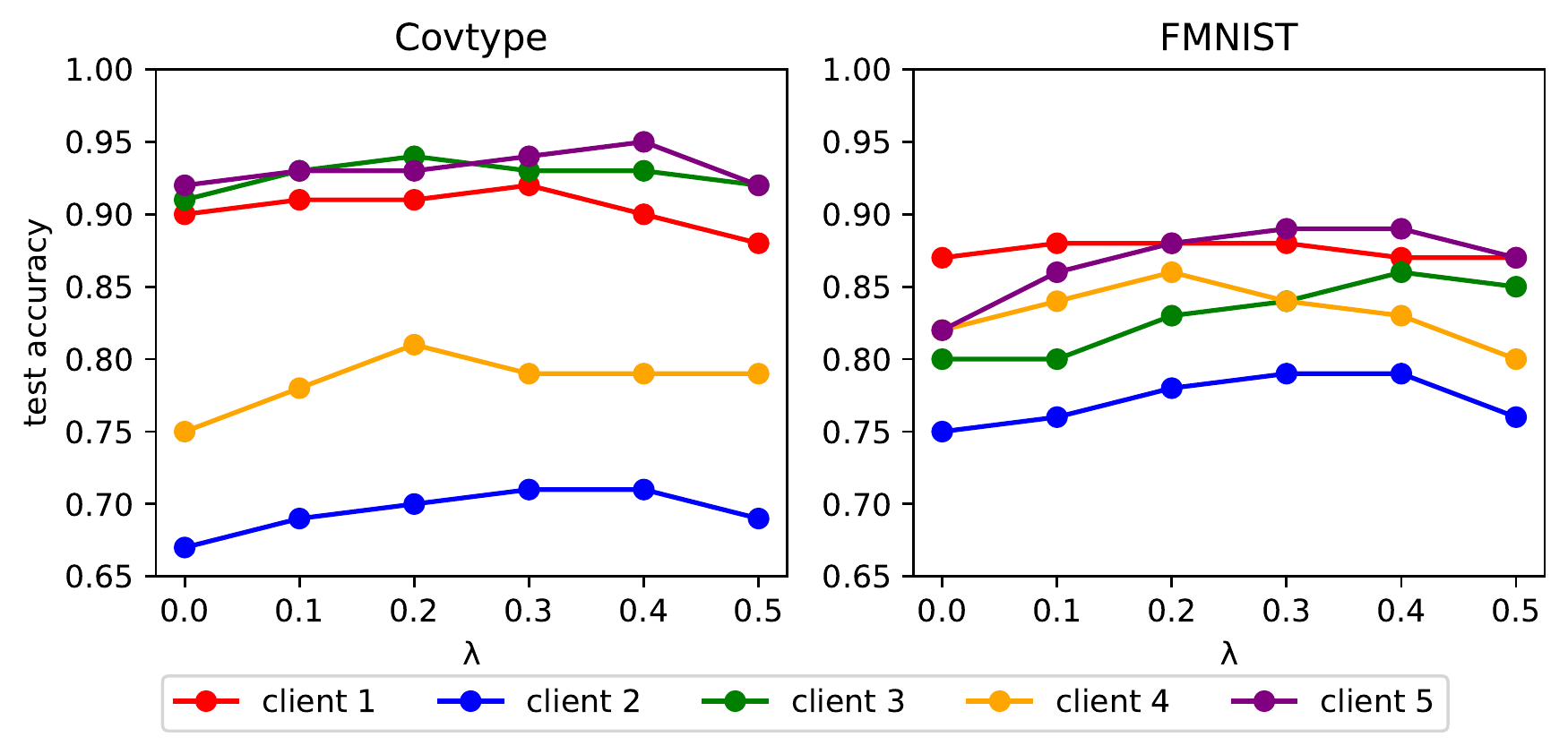}
\caption{Numerical results for the impact of the hyperparameter $\lambda$, where the x-axis represents the parameter $\lambda$ and the y-axis represents the accuracy of the test dataset (Section~\ref{sec:exp_lambda}).} 
\label{fig:exp_lambda}
\end{figure}
                                                                            
\subsection{Impact of the number of training data} \label{sec:exp_data}
In this experiment, we want to examine the performance of our approach under the setting of low data. For each client $m$, we set the hyperparameter $\lambda^m = 0.3$ and change the size of the local training set $\Dscr^m$. The result is shown in Table~\ref{tab:num_data}, where the entries represent the test accuracy. From the result, we can see that the performance of our approach improves as training data increases. Moreover, we observe that the marginal improvement decreases as training data increases for all the clients and datasets, suggesting that our proposed approach is most useful under the low data setting. 

\subsection{Impact of knowledge-based models and federated learning} \label{sec:exp_approach}
In this experiment, we want to examine the improvement brought by the KMs and the FL framework. We compare the five approaches as listed in Section~\ref{sec:case_study_numeric}. For each client, we set the hyperparameter $\lambda^m = 0.3$ and use 1\% of the training data to model the situation of low data. The result is shown in Table~\ref{tab:approach}, where entries represent the test accuracy (TA) and the percentage of violation (POV)~\eqref{eq:per_vio}. We observe that our proposed approach FLwKM consistently performs the best among the five approaches, supporting the effectiveness of our design. Besides, by comparing the performance of P-KM and MLwKM, we can see the benefit brought by the machine learning model. Similarly, by comparing the performance of FL and FLwKM, we can see the benefit brought by the KMs. 

\subsection{Impact of the hyperparameter $\lambda$} \label{sec:exp_lambda}
In this experiment, we want to see the impact of the local hyperparameter $\lambda^m$. As we illustrated in Proposition~\ref{prop:property}(b), when $\lambda^m \geq 0.5$, $f^m$ will make the same prediction as the local P-KM $g_p^m$, and when $\lambda^m = 0.0$, $f^m$ does not utilize anything from $g_p^m$. So for each client $m$, we test the performance of the model for $\lambda^m \in \{0.0, 0.1, \dots, 0.5\}$. The result is shown in Figure~\ref{fig:exp_lambda}. The curves show that a small $\lambda^m$ may not be enough to bring the benefit from $g_p^m$, and a large $\lambda^m$ may cause $g_p^m$ to dominate. The U-shape of the curves also suggests that we can use cross-validation~\citep{arlot2010survey} or bilevel optimization~\citep{bergstra2011algorithms} techniques to obtain a good $\lambda$ in practice.

\section{Conclusion}
We propose a federated learning framework where the clients can work collaboratively using local data sets and local domain knowledge without sharing such information. We show that our framework can be applied to solve a real industry-level problem. Moreover, we show that our framework can be extended to solve regression problems (Appendix C). 

There are also many interesting future directions. For example, it is interesting to extend our methodology to other types of knowledge-based models. As another example, it is also possible to extend our methodology to vertical federated learning.

\bibliographystyle{plainnat}
\bibliography{biblio.bib}

\newpage

\begin{appendices}
\section{Proof for Proposition~\ref{prop:property}} 
\begin{proof}
\begin{itemize}
        \item[(a)] Consider any $\theta\in\mR^d$ and $x\in\Xscr$. By the definition of P-KM, we know that $g_p^m(x) \in \Delta^k$. Similarly, by the definition of the $\softmax$ function, we know that 
        \[\softmax( f(\theta; x) + \infty\cdot(g_r^m(x) - \mathbf{1}^k)) \in \Delta^k.\]
        Therefore, it follows that $f^m(\theta; x) \in \Delta^k$. 
        
        \item[(b)] The first part is straightforward, namely, we have 
        \[\langle f^m(\theta; x),\ g_p^m(x)\rangle \geq \lambda^m g_p^m(x),\ g_p^m(x)\rangle = \lambda^m.\]
        Now we show the second part. Given $x\in\Xscr$, suppose that $\supp(g_p^m(x)) = \{i\} \subset [k]$. Then by the first part, it follows that for any $\theta\in\mR^d$, we have 
        \[f^m(\theta; x)_i \geq \lambda^m.\]
        Therefore, if $\lambda^m \geq 0.5$, combing Proposition~\ref{prop:property}(a), we can conclude that 
        \[\{i\} \subseteq \supp(f^m(\theta; x)).\]

        \item[(c)] By Assumption~\ref{ass:blanket}, we know that 
        \[\supp(g_p^m(x)) \subseteq \supp(g_r^m(x)) \enspace \forall x \in \Xscr.\]
        By the construction of $f^m$, we only need to show 
        \begin{equation*}
                \supp\left( \softmax( f(\theta; x) + \infty\cdot(g_r^m(x) - \mathbf{1}^k)) \right) 
            \subseteq \supp(g_r^m(x)).
        \end{equation*}
        Consider any $i \in [k] \setminus \supp(g_r^m(x))$. We have 
        \[\softmax( f(\theta; x) + \infty\cdot(g_r^m(x) - \mathbf{1}^k))_i = 0.\]
        Therefore, it follows that 
        \[i \notin \supp\left( \softmax( f(\theta; x) + \infty\cdot(g_r^m(x) - \mathbf{1}^k)) \right).\]

        \item[(d)] By the construction of $f^m$ and chain rule, we know that
        \[\nabla_\theta f^m(\theta; x) = (1-\lambda^m)\nabla_s\softmax(s)\cdot \nabla_\theta s,\]
        where 
        \[s = \softmax( f(\theta; x) + \infty\cdot(h^m\circ\Phi^m(x) - \mathbbm{1}^k) ).\]
        By the definition of $\softmax$, it can be derived that 
        \[\nabla_s\softmax(s) = \diag(s) - s^T s.\]
        Similarly, we have
        \[\nabla_\theta s = \nabla_{\theta}f(\theta; x).\]
        Finally, combing these together leads to the desired result. 
\end{itemize}
\end{proof}

\section{Mask and Maxpool operator}
In this section, we introduce two operators used in this work. 

\paragraph{Mask operator.} Let $\Omega$ be a subset of $\{1, \dots, n\}$ with $|\Omega|=k$, the mask operator $\Phi_\Omega:\mR^n\to\mR^k$ is defined as
    \begin{equation} \label{eq:mask}
        \Phi_\Omega(x)_i = x_{\Omega[i]} \enspace \forall i = 1,\dots,k,
    \end{equation}
where $\Omega[i]$ denotes the $i$-th element in $\Omega$. 

\paragraph{Maxpool operator} Given any matrix $X \in \mR^{n\times n}$, let $p$ be an integer such that $n \mod p = 0$, then the $p$-by-$p$ Maxpool operator 
    $\Phi_{(p,p)}: \mR^{n\times n} \to \mR^{k\times k}$, where $k = n / p$, 
    is defined as 
    \begin{equation} \label{eq:maxpool}
    \Phi_{(p,p)}(X)_{i,j} = \max(X_{(i-1)p+1:ip,(j-1)p+1:ip}),
    \end{equation}
where $\max(X_{(i-1)p+1:ip,(j-1)p+1:iq})$ denotes the maximal value in the submatrix.

\section{Extension to regression problem} 
So far, we have shown that our proposed approach can be applied to multiclass classification problems. This section shows that our approach can be extended to solve regression problems using a discretization technique initially developed by \citet{torgo1997regression}.

We consider a standard regression problem 
\begin{equation} \label{eq:regression}
    (x \in \Xscr, y \in [\ell, u]) \sim \Fscr,
\end{equation}
where $\Fscr$ is the underlying data distribution, $x$ is the feature vector and $y$ is the label with $\ell$ and $u$ being the finite lower and upper bounds. 

Similarly, we define the KMs for the regression problem \eqref{eq:regression}.  
\begin{definition}\label{def:reg-model}
Suppose that $g$ is a KM for the data distribution $\Fscr$ (Definition~\ref{def:know-model}).
\begin{itemize}
\item[(a)] We say that $g$ is of prediction type if it is a mapping from $\Xscr$ to $[\ell, u]$ and for any $(x,y)\sim\Fscr$, $g(x)$ is a point estimation of $y$.
\item[(b)] We say that $g$ is of range type if it is a mapping from $\Xscr$ to $2^{[\ell, u]}$ and for any $(x,y)\sim\Fscr$, one has
$$ y \in g(x). $$
\end{itemize}
\end{definition}

Now we illustrate how we apply our classification inductive
learning system on the regression problem. The main idea is to transform the regression problem into a closely related classification problem. The idea of mapping regression into classification was originally considered by \citet{weiss1995rule} and was later extensively explored by \citet{torgo1997regression}. Here we follow the discretization methodology developed in \citep{torgo1997regression}.

First, we define $k$ as the number of classes that we want to have. Next, we partition the interval $[\ell, u]$ into $k$ intervals with the same range, i.e. 
\[\Pscr_k = \left\{ [\ell, \ell + \epsilon), [\ell + \epsilon, \ell + 2\epsilon), \dots, [\ell + (k-1)\epsilon, u] \right\},\]
where $\epsilon \coloneqq \frac{u-\ell}{k}$.
Then we construct two mappings 
\[\phi_{g_p}:[\ell, u] \to \OneHot(k),\]
and
\[\phi_{g_r}:2^{[\ell, u]}\to\MultiHot(k)\]
that can transform the KMs for the regression problem (Definition~\ref{def:reg-model}) to the KMs for the classification problem (Definition~\ref{def:pred-model}). The explicit formulations of these two mappings are
\begin{equation*}
    \begin{split}
        \forall s \in [\ell, u], \enspace 
        \phi_{g_p}(s)_i &= 
        \begin{cases}
            1 & s \in [\ell + (i-1)\epsilon, \ell + i\epsilon) \\
            0 &\text{otherwise}
        \end{cases}
        \enspace\text{and}\\
        \forall S \subseteq [\ell, u], \enspace 
        \phi_{g_r}(S)_i &=
        \begin{cases}
            1 & S \cap [\ell + (i-1)\epsilon, \ell + i\epsilon) \neq \emptyset\\
            0 &\text{otherwise}.
        \end{cases}
    \end{split}
\end{equation*}

Finally, let $\{(x_i\in\Xscr, y_i\in[\ell, u])\}_{i = 1}^N$ denote the training set for the regression problem, $g_p:\Xscr\to[\ell, u]$ denote the corresponding P-KM and $g_r:\Xscr\to2^{[\ell, u]}$ denote the corresponding R-KM. Then after the transformation, we will get a classification problem with training set $\{(x_i, \phi_{g_p}(y_i))\}_{i = 1}^N$, P-KM $\phi_{g_p} \circ g_p$ and R-KM $\phi_{g_r} \circ g_r$.

\end{appendices}

\end{document}